\documentclass{article}
 \usepackage{amsmath}
 \usepackage{graphicx} 
\usepackage{amsthm}
\usepackage{thmtools}
\usepackage{wrapfig}
\usepackage{thm-restate}
\usepackage{pifont}
\usepackage{verbatim}
\newcommand{\etal}{\textit{et al.}}
\usepackage[table]{xcolor}
\definecolor{lightred}{RGB}{255, 230, 230}
\definecolor{lightgreen}{RGB}{240, 255, 240}
 \usepackage[preprint]{neurips_2026}
\usepackage[utf8]{inputenc} 
\usepackage[T1]{fontenc}    
\usepackage{hyperref}       
\usepackage{url}            
\usepackage{booktabs}       
\usepackage{amsfonts}       
\usepackage{nicefrac}       
\usepackage{microtype}      
\usepackage{xcolor}         

\newtheorem{example}{Example}
\newcommand{\blank}{\textunderscore\textunderscore\textunderscore}
\definecolor{cornellred}{rgb}{0.7, 0.11, 0.11}
\definecolor{cadmiumgreen}{rgb}{0.0, 0.42, 0.24}
\definecolor{aliceblue}{rgb}{0.91, 0.94, 0.97}
\definecolor{darkblue}{rgb}{0.83, 0.89, 0.97}
\definecolor{Red7}{rgb}{0.941, 0.243, 0.243}
\definecolor{Green7}{RGB}{55, 178, 77}
\definecolor{Blue9}{rgb}{0.098,0.3,0.9}

\hypersetup{
  linkcolor = cornellred,
  citecolor  = cadmiumgreen,
  colorlinks = true,
}

\title{Looped Diffusion Language Models}

\author{
\textbf{Sanghyun Lee}$^1$\thanks{This work was done during an internship at KRAFTON.},
\textbf{Chunsan Hong}$^1$,
\textbf{Seungryong Kim}$^1$
\\[2pt]
\textbf{Jonghyun Lee}$^2$,
\textbf{Jongho Park}$^3$,
\textbf{Dongmin Park}$^{2}$\thanks{Corresponding author.}
\\
$^1$KAIST
\quad
$^2$KRAFTON
\quad
$^3$University of California, Berkeley
\\
\texttt{\{lsh83210, hoarer, seungryong.kim\}@kaist.ac.kr}
\\
\texttt{\{jonghyunlee, dongmin.park\}@krafton.com}
\quad
\texttt{jjhpark@berkeley.edu}
}

\begin{document}

\maketitle

\begin{abstract}
Masked diffusion models (MDMs) have emerged as a promising alternative to autoregressive models for language modeling, yet the effective design of transformer architectures for MDMs remains underexplored.
In this paper, we show that selectively looping the early-middle transformer layers significantly improves both training efficiency and model performance in MDMs.
We call this approach \textbf{LoopMDM} (Looped Masked Diffusion Model), which brings two key benefits: looping layers at training-time yields a depth-scaling effect without adding parameters, while varying the number of loops at inference-time enables flexible compute scaling.
Despite the simplicity, the results are striking:
across multiple pre-training corpora, LoopMDM matches the performance of same-size MDMs with up to $3.3\times$ fewer training FLOPs, while its final performance outperforms them on various reasoning benchmarks, including up to $+8.5$ points on GSM8K.
It even surpasses deeper non-looped MDMs trained with comparable per-step compute, indicating that selective looping is more effective than naive depth scaling.
Furthermore, LoopMDM can scale inference-time compute by increasing the number of loops. 
Adaptively adjusting the number of loops throughout the sampling process further yields additional gains in compute efficiency while maintaining performance.
Lastly, with attention analysis, we provide evidence that looping is effective in MDMs by promoting interactions among masked positions.
\end{abstract}

\section{Introduction}

Masked diffusion models (MDMs)~\citep{austin2021structured, lou2024discrete, sahoo2024simple, shi2025simplified} have emerged as a competitive alternative to autoregressive models (ARMs) for language modeling. Recent advances in training objectives~\citep{peng2025planner}, noise schedules~\citep{hong2026unifying}, and architectures~\citep{chao2025beyond, deschenaux2025partition} have steadily narrowed the gap with ARMs, establishing MDMs as an increasingly viable framework for text generation. Building on these improvements, recent work has further pushed MDMs toward larger-scale models, including the LLaDA family~\citep{nie2025llada,zhu2025llada,bie2025llada2}, Dream~\citep{ye2025dream}, and Seed Diffusion~\citep{song2025seed}, to narrow the gap even further. Although these approaches demonstrate the promise of scaling, they also raise the question of how to improve performance along axes other than parameters and training tokens.

A natural candidate from the AR literature is the \emph{looped transformer}~\citep{dehghani2018universal, giannou2023looped, geiping2025scaling, zhu2025scaling}. 
Looped architectures apply a shared block repeatedly to convert depth into loops at a fixed parameter cost. Looping has been explored in ARMs as a form of test-time compute scaling, but its role in MDMs has not been explored. 
It is unclear whether looping should help in this setting since MDMs already perform iterative computation across denoising steps. Moreover, previous works on looping have focused on parameter efficiency~\citep{saunshi2025reasoning} or inference-time refinement~\citep{yang2023looped,fan2024looped},
leaving much to be desired regarding training-compute efficiency, as performance gain over non-looped models at fixed compute remains non-trivial~\citep{schwethelm2026much}.

\begin{figure}
    \centering
    \vspace{-0.5em}
    \includegraphics[width=1\linewidth]{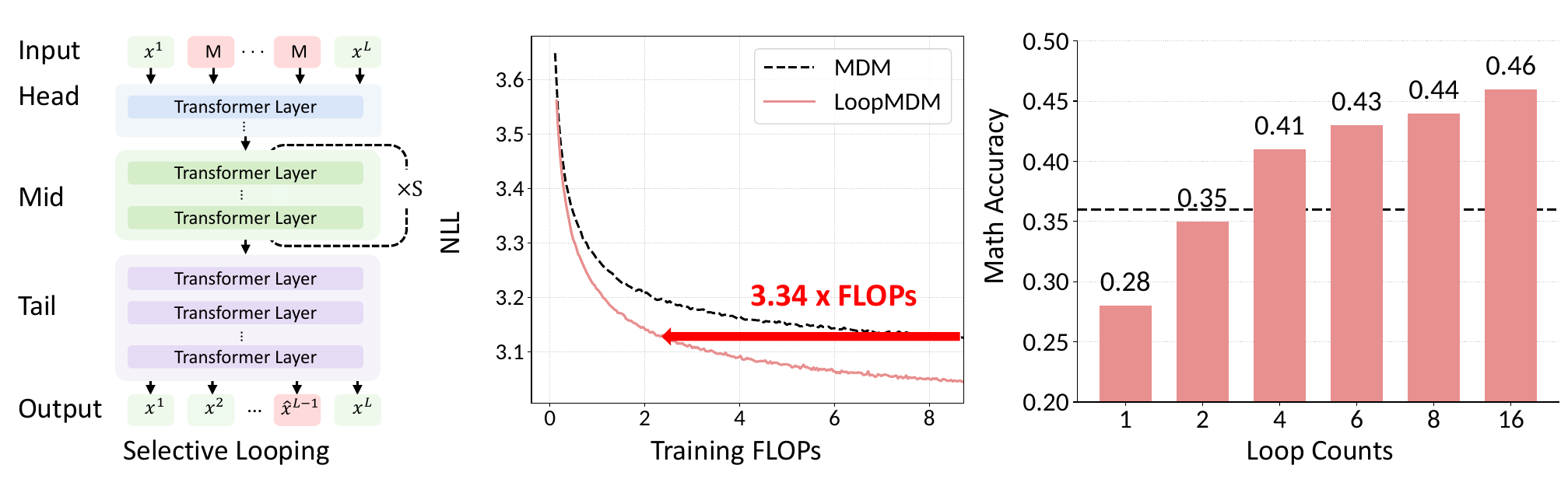}
    \vspace{-1.0em}
\caption{
\textbf{Overview of LoopMDM.}
(Left) LoopMDM selectively applies looping to a small early-middle layers of the denoising network. 
(Middle) Under matched training compute, LoopMDM (red) reaches the same test NLL as a non-looped MDM baseline with the same architecture (dashed black) using substantially fewer training FLOPs. 
(Right) Increasing the inference-time loop count consistently improves GSM8K accuracy; the dashed line denotes the same non-looped MDM baseline.}
    \label{fig:overview}
    \vspace{-1.3em}
\end{figure}

Motivated by this gap, we study how to integrate looping into MDMs under matched training compute. Through a systematic study of where and how much to loop, we find that selectively looping a few early-middle transformer layers captures the full benefit of repeated computation at substantially lower training cost than uniform looping. We refer to this configuration as \textbf{LoopMDM}, the first application of looped transformers to masked diffusion language models. Figure~\ref{fig:overview} provides an overview of the architecture and its scaling behavior. Three observations summarize our findings:

\paragraph{Training compute efficiency.} Across the three pretraining corpora, including FineWeb-Edu~\citep{penedo2024fineweb}, OpenWebText~\citep{Gokaslan2019OpenWeb}, and LM1B~\citep{chelba2013lm1b}, LoopMDM matches the test negative log-likelihood (NLL) of an equal-size MDM with up to $3.3\times$ fewer training floating-point operations (FLOPs). When trained at matched total training FLOPs, it consistently outperforms the baseline over various benchmarks in zero-shot perplexity, generative perplexity, and downstream task accuracy. Crucially, the observed efficiency gains depend strongly on \textit{how} looping is applied, with a small number of early-middle layers already capturing most of the benefit, motivating the selective design.

\paragraph{Gains beyond depth scaling.} On GSM8K, LoopMDM improves its same-size non-looping MDM baseline by $+8.5\%$, and also outperforms deeper MDMs whose per-step training FLOPs match those of LoopMDM. This suggests that the gains from looping are not merely a proxy for additional depth, but arise from the repeated application of a shared computation, particularly in math tasks.

\paragraph{Insights into looping and adaptive inference.} We analyze how looping interacts with masked positions through a sequence of controlled studies. A Sudoku experiment with a restricted generation order shows that looping enables the model to solve a globally constrained task that a shallow baseline cannot solve. Attention analysis further reveals that mask-to-mask interactions increase with loop counts, and a timestep-level study shows that these gains are concentrated at intermediate noise levels. These observations motivate a simple adaptive strategy that allocates more loop counts where they are most effective, retaining performance while reducing compute.

Together, our findings suggest that masked diffusion language models have significant room to improve their training efficiency and performance through architectural or modeling choices, evidenced by our selective looping approach, and we hope this work sparks further exploration in this direction.

\section{Background}
We briefly review masked diffusion language models and the training objective used throughout this work. Related work is discussed in Appendix~\ref{sec:related}.
\subsection{Masked Diffusion Models}
\paragraph{Notation.}
We consider sequences of length $L$, denoted $x = (x^1, x^2, \ldots, x^L)$, where each token $x^i$ takes values in a vocabulary $\mathcal{V} = \{1, \ldots, V\}$ and is represented as a one-hot vector. We write $\bar{\mathcal{V}} = \mathcal{V} \cup \{\mathrm{m}\}$ for the vocabulary augmented with a mask token $\mathrm{m}$, and use $\mathrm{Cat}(\cdot \mid p)$ to denote a categorical distribution parameterized by probabilities $p$ over the vocabulary.

\paragraph{Forward process.}
Let $\alpha_t : [0,1] \to [0,1]$ be a strictly decreasing schedule with $\alpha_0 \simeq 1$ and $\alpha_1 \simeq 0$.  We use $\delta_a(b)$ to denote the point mass at $a$. The forward process independently corrupts each token of a clean sequence $x_0 \sim p_{\text{data}}$ via
\begin{equation}
    q(x_t^i \mid x_0^i) = \alpha_t\,\delta_{x_0^i}(x_t^i) + (1 - \alpha_t)\,\delta_{\mathrm{m}}(x_t^i).
\end{equation}

\paragraph{Reverse process.}
The reverse process assumes conditional independence across positions. Starting from $x_1 = (\mathrm{m}, \ldots, \mathrm{m})$, a denoising network $p_\theta(x_0^i \mid x_t)$ predicts the clean-token distribution at every masked position. Letting $\alpha_{s|t} = (1-\alpha_s) / (1-\alpha_t)$, the reverse transition for $0 \leq s < t$ keeps the unmasked tokens intact and updates each masked position as
\begin{equation}
    p_\theta(x_s^i \mid x_t) = \alpha_{s|t}\,\delta_m(x_t^i) + (1 - \alpha_{s|t})\,\mathrm{Cat}(x_s^i \mid p_\theta(x_0^i \mid x_t)), \quad x_t^i = \mathrm{m}.
\end{equation}

The model is trained by minimizing the negative evidence lower bound (NELBO):
\begin{equation}
    \mathcal{L}_{\text{NELBO}} = \mathbb{E}_{t \sim \mathcal{U}[0,1],\, x_t \sim q(x_t \mid x_0)} \left[\sum_{i=1}^{L} \mathbb{I}[x_t^i = \mathrm{m}]\, \frac{\alpha'_t}{1 - \alpha_t} \log \langle p_\theta^i(x_t, t),\, x_0^i \rangle \right],
\end{equation}
where $\mathbb{I}[\cdot]$ denotes the indicator function and $\langle \cdot, \cdot \rangle$ the standard inner product.
\section{Looped Masked Diffusion Models}
We introduce \emph{Looped Masked Diffusion Model} 
(LoopMDM), an architecture that integrates looping 
into the denoising network of a masked diffusion model. The 
design is guided by two principles: (i) loop counts should 
be applied selectively to layers where iterative refinement 
yields the largest benefit per unit of training FLOPs, rather 
than uniformly across the network; and (ii) the model should 
support effective scaling with the number of loop counts, 
producing monotonic gains as loop counts $S$ increases at inference time. 
We describe each component below.

\paragraph{Architecture.}
For exposition, we divide the denoising transformer into three sequential groups of layers, a \texttt{head} of $n_h$ layers, a \texttt{mid-block} of $n_m$ layers, and a \texttt{tail} of $n_o$ layers, where the tail includes an LM head that maps the final hidden states to per-position token distributions. The head and tail are applied once per forward pass, while the mid-block is a \emph{looped} stack whose weights are shared across $S$ successive applications. We keep the head and tail unshared to preserve their specialization for input embedding and output projection.

Given an input $x_t \in \bar{\mathcal{V}}^L$, the head produces an initial hidden state $H_t^{(0)} \in \mathbb{R}^{L \times d}$, the mid-block is applied $S$ times reusing the same parameters, and the tail produces $p^S_\theta(x_0 \mid x_t)$:
\begin{equation}
    H_t^{(0)} = \texttt{head}(x_t), \quad
    H_t^{(k)} = \texttt{mid}\!\left(H_t^{(k-1)}\right), \quad
    p^S_\theta(x_0 \mid x_t) = \texttt{tail}(H_t^{(S)}).
\end{equation}
 The forward pass unrolls to an effective depth of $n_h + S \cdot n_m + n_o$ layers, while the number of parameters matches a non-looped network with $n_h + n_m + n_o$ layers.

\paragraph{Configuring the looped block.}
Three factors govern the compute–accuracy trade-off: the number of looped layers $n_m$, the position of the looped block, and the maximum loop $S_{\text{max}}$ during training. We find that effective performance does not require looping the entire network. Instead, selectively looping a small number of layers is sufficient to capture most of the gains, while looping more layers increases training cost without further benefit.

The placement of the loop is equally important. Looping early-middle layers consistently yields the strongest results, suggesting that repeated application is most effective after low-level token representations have formed, but before later layers become specialized for final prediction. Similar observations have been reported in recent studies of looped transformers, where intermediate layers provide a favorable balance between representation stability and iterative computation~\citep{bae2025mixture, saunshi2025reasoning}. A detailed ablation of these factors is presented in Section~\ref{sec:ablation}.

\paragraph{Training with stochastic loop count.}
We train LoopMDM by minimizing the standard MDM NELBO 
applied to the final loop output:
\begin{equation}
    \mathcal{L}_{\text{NELBO}} = 
    \mathbb{E}_{t,\, S,\, x_t} 
    \left[\sum_{i=1}^{L} \mathbb{I}[x_t^i = \mathrm{m}]\, 
    \frac{\alpha'_t}{1 - \alpha_t} 
    \log \langle p_\theta^{S,i}(x_t, t),\, x_0^i \rangle \right],
\end{equation}
where $t \sim \mathcal{U}[0,1]$ is sampled uniformly over timesteps, 
$x_t \sim q(x_t \mid x_0)$, 
and the loop count $S \sim \mathcal{U}\{1, \ldots, S_{\max}\}$ is sampled uniformly from the discrete set $\{1, \ldots, S_{\max}\}$ at each training step. 
This stochastic schedule exposes the shared mid-block to a range of effective depths, yielding representations that remain stable across iteration counts and enabling deployment with any $S$ at inference, including values beyond $S_{\max}$.

\paragraph{Training cost.}
Following standard convention~\citep{kaplan2020scaling}, the training FLOPs of a transformer can be approximated as $C = 6ND$, where $N$ is the number of parameters and $D$ is the number of training tokens. Looped architectures differ in that, while the parameter count remains unchanged, looped layers are executed multiple times per forward pass, increasing the per-step compute. For a model with $n_m$ mid layers and maximum loop count $S_{\max}$, the per-step FLOPs of LoopMDM are
\begin{equation}
    F_{\text{loop}} = F_{\text{base}} 
    + \left(\mathbb{E}[S] - 1\right) \cdot n_m \cdot F_{\text{layer}},
\end{equation}
where $\mathbb{E}[S] = (S_{\max}+1)/2$ under stochastic sampling, $F_{\text{base}}$ is the per-step FLOPs of the baseline, and $F_{\text{layer}}$ is the FLOPs of a single transformer layer. 

To ensure a fair comparison, we match \emph{total training FLOPs} by training LoopMDM for proportionally fewer steps, so that all reported gains arise from architectural differences, not additional compute.

\section{Experiments}

We organize this section around three findings: (i) selective looping significantly reduces training compute while improving multiple quality metrics ($\S$ \ref{sec:finding-efficiency}), (ii) the gains are most pronounced on math reasoning tasks, where LoopMDM outperforms deeper MDMs with comparable per-step compute ($\S$ \ref{sec:finding-math}), and (iii) looping allows for increased interactions among masked positions and exhibits timestep-dependent behavior, which together motivate a simple adaptive inference strategy ($\S$ \ref{sec:insight}). 

\subsection{Selective looping matches MDM quality with substantially fewer training FLOPs}
\label{sec:finding-efficiency}

\begin{figure}
    \centering
    \includegraphics[width=\textwidth]{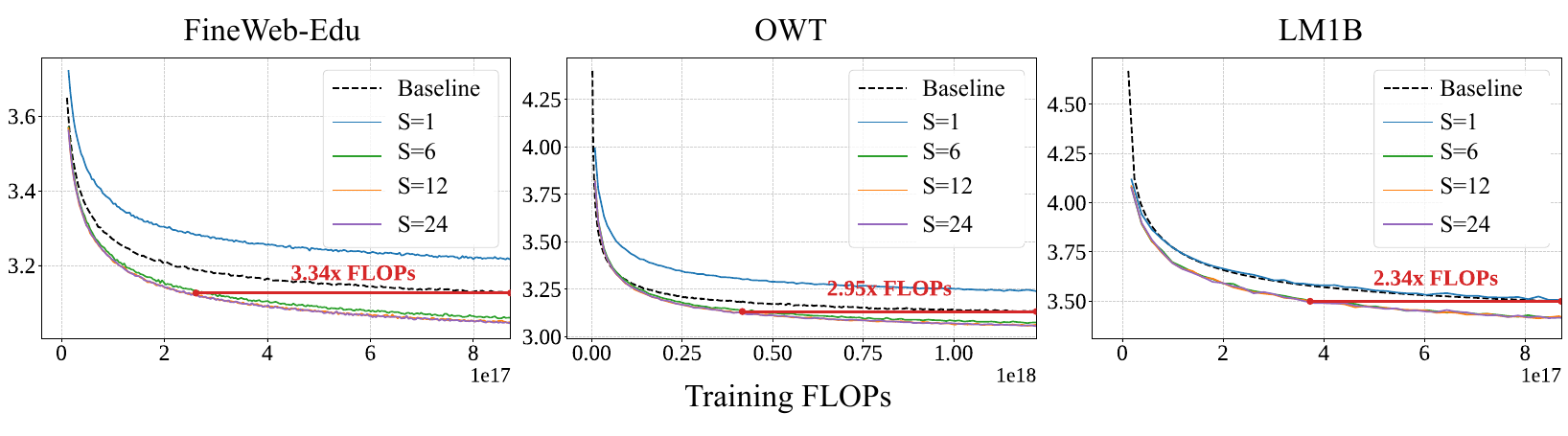}
    \vspace{-0.8cm}
    \caption{\textbf{Test NLL across language pre-training datasets.}
Test NLL as a function of training FLOPs on FineWeb-Edu, OpenWebText (OWT), and LM1B. 
All models are iso-parameter (170M) and trained under matched training FLOPs. 
Looping is applied to two mid-layers 1-2 in zero-based indexing. Solid curves show LoopMDM with varying inference-time loop counts ($S=1,6,12,24$), where $S=24$ exceeds the maximum loop count used during training ($S_{\max} = 12$); the dashed curve shows the MDM baseline. Performance improves consistently with increasing $S$.}
    \label{fig:test}
\end{figure}

Our experiments show that the gains of LoopMDM depend critically on selective loop design. By applying looping only to a small set of early-middle layers, LoopMDM achieves performance comparable to or better than the MDM baseline while using only a fraction of the original training compute. Under this selective configuration, LoopMDM matches the baseline test NLL with up to $3.3\times$ fewer training FLOPs (Figure~\ref{fig:test}), and consistently improves zero-shot perplexity, generative perplexity, and downstream performance across all evaluation settings (Tables~\ref{tab:genppl}--\ref{tab:downstream}).

\paragraph{Experimental setup.}
We train two models, LoopMDM and a non-looped MDM baseline, matched in parameter count, training data, and total training FLOPs, differing only in per-step training and inference compute, which are controlled by the loop count $S$. We train both models on three pretraining corpora, including OpenWebText~\citep{Gokaslan2019OpenWeb}, LM1B~\citep{chelba2013lm1b}, FineWeb-Edu~\citep{penedo2024fineweb}, using a 170M-parameter diffusion transformer with rotary positional encoding and adaptive layer normalization, following prior work~\citep{sahoo2024simple, shi2025simplified}. Sequence length is 128 for LM1B, and 1024 for OpenWebText and FineWeb-Edu. The MDM baseline is trained for 1M steps with a batch size of 512 and sequence packing, while 12-layer LoopMDM loops two (first and second) mid-layers 1-2 in zero-based indexing with $S_{\max}=12$, without increasing the parameter count. For downstream evaluation, we use FineWeb-Edu checkpoints under matched training compute and follow the protocol of~\citep{sahoo2024simple, nie2024scaling}, which are detailed in Appendix~\ref{appendix:expdetail}.

\paragraph{Test NLLs.}
Figure~\ref{fig:test} reports test negative log-likelihood (NLL) on LM1B, OpenWebText, and FineWeb-Edu. For $S \geq 6$, LoopMDM consistently outperforms the MDM baseline across all three corpora, reaching the same final NLL with $3.34\times$, $2.95\times$, and $2.34\times$ fewer training FLOPs, respectively. The behavior at extreme loop counts $S$ further clarifies this trend. At $S=1$, LoopMDM underperforms the baseline, consistent with prior observations that shared blocks behave as looping operators whose benefits emerge only through repeated application~\citep{dehghani2018universal, geiping2025scaling, zhu2025scaling}. At $S=24$, beyond the training maximum $S_{\max}=12$, performance continues to improve, indicating that the learned operator generalizes beyond the training regime.

\begin{table}[t!]
    \small
    \centering
    \caption{\textbf{Generative perplexity ($\downarrow$) across 
    sampling steps.} $S$ denotes the inference-time loop count. 
    Generative perplexity decreases monotonically with $S$ across 
    all denoising step budgets. Results are averaged over five random seeds.}
    \begin{tabular}{lccccc}
        \toprule
        Steps & 64 & 128 & 256 & 512 & 1024 \\
        \midrule
        MDM & 116.71{\scriptsize$\pm$ 2.72} & 84.71 {\scriptsize$\pm$ 2.15} & 79.43{\scriptsize$\pm$ 2.60} & 55.50{\scriptsize$\pm$ 2.04} & 42.56{\scriptsize$\pm$ 1.09} \\
        $S=1$ & 116.74{\scriptsize$\pm$ 2.66}  & 82.71{\scriptsize$\pm$ 2.35} & 67.39 {\scriptsize$\pm$ 1.86}& 53.13{\scriptsize$\pm$ 0.78} & 43.86{\scriptsize$\pm$ 1.39} \\
        $S=6$ & 105.59{\scriptsize$\pm$ 3.83} & 76.76{\scriptsize$\pm$ 1.09} & 64.08{\scriptsize$\pm$ 1.39} & 49.16{\scriptsize$\pm$ 2.65} & 39.23{\scriptsize$\pm$ 1.31} \\
        $S=12$ & \textbf{99.06{\scriptsize$\pm$ 5.91}} & \textbf{68.22{\scriptsize$\pm$ 1.58}} & \textbf{56.45}{\scriptsize$\pm$ 0.50} & \textbf{48.16{\scriptsize$\pm$ 3.17}} & \textbf{38.53{\scriptsize$\pm$ 1.23}} \\
        \bottomrule
    \end{tabular}
    \vspace{-0cm}
    \label{tab:genppl}
\end{table}

\begin{table}[t!]
    \centering
    \small
    \caption{\textbf{Zero-shot perplexities ($\downarrow$).}
    Performance tends to improve with increasing loop counts $S$. Results are averaged over five random seeds.}

    \setlength{\tabcolsep}{2.7pt}
    \label{tab:zeroshot-ppl}
    {
    \begin{tabular}{lccccccc}
    \toprule
    \textbf{Method} & \multicolumn{7}{c}{\textbf{Evaluation benchmark}} \\
    \cmidrule(l){2-8}
     & PTB & WikiText & LM1B & Lambada & AG News & PubMed & Arxiv \\
    \midrule
    MDM & 108.5{\scriptsize$\pm$ 6.61}
    & 35.1 {\scriptsize$\pm$ 1.11}
    & 68.1 {\scriptsize$\pm$ 2.37}
    & 49.6 {\scriptsize$\pm$ 1.67}
    & 69.3 {\scriptsize$\pm$ 1.96}
    & 43.0 {\scriptsize$\pm$ 0.42}
    & 37.8{\scriptsize$\pm$1.01 } \\
    
    $S=1$    
    & 124.6 {\scriptsize$\pm$ 12.02}
    & 41.8 {\scriptsize$\pm$ 1.54}
    & 77.4 {\scriptsize$\pm$ 2.69}
    & 53.3 {\scriptsize$\pm$ 2.10}
    & 87.8 {\scriptsize$\pm$ 2.17}
    & 47.9 {\scriptsize$\pm$ 0.71}
    & 42.5 {\scriptsize$\pm$ 1.49} \\
    
    $S=6$    
    & 100.2 {\scriptsize$\pm$ 9.44}
    & 34.2 {\scriptsize$\pm$ 1.29}
    & 68.0 {\scriptsize$\pm$ 2.72}
    & \textbf{44.6} {\scriptsize$\pm$ 1.91}
    & 59.2 {\scriptsize$\pm$ 1.86}
    & 37.2 {\scriptsize$\pm$ 0.71}
    & 33.1 {\scriptsize$\pm$ 1.01} \\
    
    $S=12$   
    & \textbf{90.2} {\scriptsize$\pm$ 8.59}
    & \textbf{32.4} {\scriptsize$\pm$ 1.31}
    & \textbf{67.8} {\scriptsize$\pm$ 2.78}
    & 45.8 {\scriptsize$\pm$ 1.90}
    & \textbf{56.6} {\scriptsize$\pm$ 1.70}
    & \textbf{36.5} {\scriptsize$\pm$ 0.70}
    & \textbf{32.7} {\scriptsize$\pm$ 0.97} \\
    \bottomrule
    \end{tabular}
    \vspace{-1em}
    }
\end{table}

\paragraph{Sampling quality.}
Table~\ref{tab:genppl} reports \textit{generative perplexity} for unconditionally sampled sequences across varying numbers of denoising steps. LoopMDM consistently improves over the baseline, achieving the lowest perplexity at $S_{\max}=12$, across all step budgets. The gains are more pronounced at smaller step counts, indicating that looping is particularly effective when the denoising process is limited. Qualitative samples are provided in Appendix~\ref{appendix:qual}.

\paragraph{Zero-shot perplexities.} Following prior work~\citep{sahoo2024simple}, we measure \emph{zero-shot perplexity} on seven external corpora unseen during training, including the validation splits of Penn Tree Bank (PTB)~\citep{marcus1993building}, WikiText~\citep{merity2016pointer}, LM1B~\citep{chelba2013lm1b}, Lambada~\citep{paperno2016lambada}, AG News~\citep{zhang2015character}, and the scientific papers from Scientific Papers (Pubmed and Arxiv;~\citep{cohan2018discourse}).
Table~\ref{tab:zeroshot-ppl} reports zero-shot perplexity on seven external corpora. LoopMDM consistently improves over the baseline across all datasets, achieving gains of $0.3$-$18.3$ points at $S=12$, with the largest improvement on PTB ($108.5 \to 90.2$). Already at $S=6$, it yields substantial improvements, indicating that most gains are realized with moderate loop counts.

\begin{table}[h]
    \centering
    \caption{\textbf{Downstream task accuracy (\%) across models.}
Performance improves consistently with increasing $S$, with looped models outperforming the MDM baseline on most benchmarks. Results are averaged over five random seeds.
}
    \resizebox{\textwidth}{!}{%
    \begin{tabular}{lccccccccc|c}
        \toprule
        Method & ARC-c & ARC-e & BoolQ & HellaSwag & OBQA & PIQA & RACE & SIQA & WinoGrande & Avg. \\
        \midrule
        MDM 
        & 25.5{\scriptsize$\pm$ 0.7}
        & 47.7{\scriptsize$\pm$ 0.9}
        & 46.6{\scriptsize$\pm$ 1.1}
        & 31.9{\scriptsize$\pm$ 0.3}
        & 30.8{\scriptsize$\pm$ 1.3}
        & 60.6{\scriptsize$\pm$ 0.2}
        & 28.5{\scriptsize$\pm$ 0.8}
        & 36.8{\scriptsize$\pm$ 0.4}
        & 51.3{\scriptsize$\pm$ 0.7}
        & 40.0{\scriptsize$\pm$ 0.3} \\

        $S=1$ 
        & 23.6{\scriptsize$\pm$ 0.8}
        & 45.2{\scriptsize$\pm$ 0.6}
        & 53.8{\scriptsize$\pm$ 0.8}
        & 31.1{\scriptsize$\pm$ 0.5}
        &  \textbf{28.2}{\scriptsize$\pm$ 0.6}
        & 58.5{\scriptsize$\pm$ 0.6}
        & 27.2{\scriptsize$\pm$ 0.7}
        & 35.9{\scriptsize$\pm$ 0.6}
        & 50.9{\scriptsize$\pm$ 0.5}
        & 39.4{\scriptsize$\pm$ 0.3} \\

        $S=6$ 
        & \textbf{25.7}{\scriptsize$\pm$ 0.2}
        &  \textbf{47.7}{\scriptsize$\pm$ 1.2}
        & 53.8{\scriptsize$\pm$ 1.8}
        & 32.5{\scriptsize$\pm$ 0.4}
        & 28.0{\scriptsize$\pm$ 0.2}
        & 58.9{\scriptsize$\pm$ 0.6}
        & 31.3{\scriptsize$\pm$ 0.5}
        &  \textbf{37.4}{\scriptsize$\pm$ 0.3}
        & 51.4{\scriptsize$\pm$ 0.3}
        & 40.7{\scriptsize$\pm$ 0.9} \\

        $S=12$ 
        & 25.3{\scriptsize$\pm$ 0.9}
        &  \textbf{47.7}{\scriptsize$\pm$ 1.6}
        &  \textbf{57.4}{\scriptsize$\pm$ 1.3}
        & \textbf{ 32.7}{\scriptsize$\pm$ 1.2}
        & 28.0{\scriptsize$\pm$ 0.6}
        &  \textbf{61.1}{\scriptsize$\pm$ 0.5}
        &  \textbf{31.7}{\scriptsize$\pm$ 0.4}
        & 37.3{\scriptsize$\pm$ 0.8}
        &  \textbf{51.5}{\scriptsize$\pm$ 0.8}
        & \textbf{41.4}{\scriptsize$\pm$ 0.6} \\
        
        \bottomrule
    \end{tabular}%
    }
    
    \label{tab:downstream}
\end{table}
\paragraph{Downstream tasks.} We evaluate \emph{zero-shot downstream accuracy} with 5 random seeds on nine widely used benchmarks spanning three categories: {commonsense reasoning} (HellaSwag~\citep{zellers2019hellaswag}, PIQA~\citep{bisk2020piqa}, Social IQA(SIQA)~\citep{sap2019social}, WinoGrande~\citep{sakaguchi2021winogrande}), {reading comprehension} (BoolQ~\citep{clark2019boolq}, RACE~\citep{lai2017race}), and {scientific question answering} (OpenBookQA (OBQA)~\citep{mihaylov2018can}, ARC-Easy (ARC-e), ARC-Challenge (ARC-c)~\citep{clark2018think}).

Table~\ref{tab:downstream} reports zero-shot accuracy on nine benchmarks. LoopMDM at $S=12$ matches or improves over the baseline on all nine, with the largest gains on BoolQ ($+10.8$) and RACE ($+3.2$). 
Interestingly, despite LoopMDM seeing fewer training tokens under matched total training FLOPs, LoopMDM matches or improves the baseline on downstream tasks.
One might have expected fewer training tokens to translate into reduced knowledge for tasks that primarily probe stored facts, yet this is not what we observe. Instead, the per-step compute spent on looping appears to compensate for the reduced token exposure, leaving downstream knowledge transfer intact.

\paragraph{Gains arise from selectively looping early-middle layers.}
\label{sec:ablation}
\begin{table}[t]
    \centering
    \small
    \caption{\textbf{Ablation on loop design (test NLL $\downarrow$).}
We vary loop position, number of looped layers $n_m$, and maximum loop count $S_{\max}$ under matched training FLOPs. 
Looping is most effective when applied to mid-layers 1-2 in zero-based indexing, with a small number of looped layers, and moderate loop counts. 
Performance degrades at both extremes, showing naive or excessive looping is inefficient. 
Green/red indicates improvements/degradations relative to the non-looped MDM baseline (3.798).}
    \label{tab:ablation}
    \begin{tabular}{cccccccccc}
        \toprule
        \multicolumn{4}{c}{\textbf{(a) Loop position}} 
        & \multicolumn{4}{c}{\textbf{(b) \# Looped layers ($n_m$)}} 
        & \multicolumn{2}{c}{\textbf{(c) Max loop ($S_{\max}$)}} \\
        \cmidrule(lr){1-4} \cmidrule(lr){5-8} \cmidrule(lr){9-10}
        Position & NLL & Position & NLL & \# Layers & NLL & \# Layers & NLL & $S_{\max}$ & NLL \\
        \midrule
        \cellcolor{lightgreen}0 & \cellcolor{lightgreen}3.744 & \cellcolor{lightgreen}6  & \cellcolor{lightgreen}3.735 & \cellcolor{lightgreen}1 & \cellcolor{lightgreen}3.743 & \cellcolor{lightred}7  & \cellcolor{lightred}3.857 & \cellcolor{lightgreen}4  & \cellcolor{lightgreen}3.741 \\
        \cellcolor{lightgreen}1 & \cellcolor{lightgreen}\textbf{3.729} & \cellcolor{lightgreen}7  & \cellcolor{lightgreen}3.739 & \cellcolor{lightgreen}2 & \cellcolor{lightgreen}\textbf{3.729} & \cellcolor{lightred}8  & \cellcolor{lightred}3.825 & \cellcolor{lightgreen}8  & \cellcolor{lightgreen}3.733 \\
        \cellcolor{lightgreen}2 & \cellcolor{lightgreen}\textbf{3.729} & \cellcolor{lightgreen}8  & \cellcolor{lightgreen}3.747 & \cellcolor{lightgreen}3 & \cellcolor{lightgreen}3.734 & \cellcolor{lightred}9 & \cellcolor{lightred}3.842 & \cellcolor{lightgreen}12 & \cellcolor{lightgreen}\textbf{3.729} \\
        \cellcolor{lightgreen}3 & \cellcolor{lightgreen}3.733 & \cellcolor{lightgreen}9  & \cellcolor{lightgreen}3.756 & \cellcolor{lightgreen}4 & \cellcolor{lightgreen}3.793 & \cellcolor{lightred}10 & \cellcolor{lightred}3.854 & \cellcolor{lightgreen}16 & \cellcolor{lightgreen}3.740 \\
        \cellcolor{lightgreen}4 & \cellcolor{lightgreen}3.734 & \cellcolor{lightgreen}10 & \cellcolor{lightgreen}3.777 & \cellcolor{lightgreen}5 & \cellcolor{lightgreen}3.794 & \cellcolor{lightred}11 & \cellcolor{lightred}3.881  & \cellcolor{lightgreen}20 & \cellcolor{lightgreen}3.763  \\
        \cellcolor{lightgreen}5 & \cellcolor{lightgreen}3.734 & -- & -- & \cellcolor{lightgreen}6 & \cellcolor{lightgreen}3.784 & \cellcolor{lightred}12 & \cellcolor{lightred}3.933 & \cellcolor{lightred}24 & \cellcolor{lightred}3.800 \\
        \bottomrule
    \end{tabular}
    \vspace{-0.3cm}
\end{table}
The performance gains above depend critically on how looping is applied. Under a controlled ablation setting on LM1B using a 170M-parameter model trained for 0.1M steps under matched training FLOPs, Table~\ref{tab:ablation} shows that selectively looping a small number of early-middle layers yields the strongest results, while broader or poorly placed looping can underperform the non-looped baseline.

Table~\ref{tab:ablation}(a) shows that loop position is highly consequential. Applying looping to early-middle layers (positions 1 or 2 in zero-based indexing) yields the best performance, with both positions achieving the same minimum NLL. In contrast, placing the loop near the input or output layers consistently degrades performance. This suggests that repeated application of the shared block is most effective at early-middle stages of the network, rather than directly on embedding or final prediction layers.

Table~\ref{tab:ablation}(b) further shows that increasing the number of mid layers is inefficient. A small number of looped layers already captures most of the gains, while larger values of $n_m$ consistently degrade performance under the same compute budget. In particular, looping more than six layers underperforms the baseline, indicating that simply applying looping to more layers does not lead to more effective use of training FLOPs.

Finally, Table~\ref{tab:ablation}(c) shows that the maximum loop count exhibits a similar trade-off. Small values of $S_{\max}$ limit the benefits of repeated application, while excessively large values degrade performance, with moderate loop counts performing best. Together, these results show that selective application is a critical component of effective training.

\subsection{Looping outperforms deeper MDMs at matched per-step training FLOPs in math}
\label{sec:finding-math}
\begin{figure}
    \centering
    \hspace{-0.5cm}\includegraphics[width=0.8\textwidth]{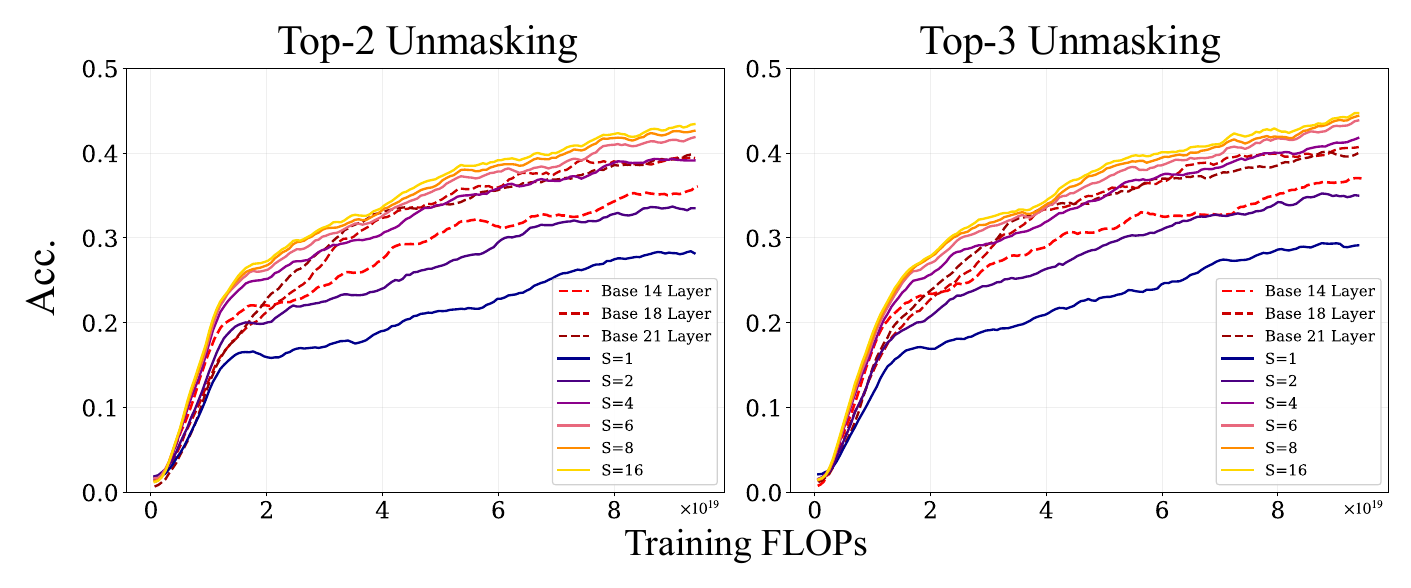}
    \vspace{-0.3cm}
    \caption{
    \textbf{GSM8K accuracy as a function of training FLOPs.} 
    14 layers LoopMDM (solid) is compared against MDM baselines with 
    14, 18, and 21 layers (dashed); the 21-layer baseline is sized 
    so that its per-step training FLOPs approximately match those of LoopMDM. 
    Inference-time loop counts $S \in \{1, 2, 4, 6, 8, 16\}$ are 
    shown, with $S\!=\!16$ exceeding the training maximum. 
    Results are reported under both Top-2 (left) and Top-3 
    (right) unmasking rules.}
    \label{fig:gsm8k}
    \vspace{-0.3cm}
\end{figure}

Our experiments on GSM8K~\citep{cobbe2021gsm8k} show that LoopMDM improves over a same-size baseline by up to $+8.5$ points while matching its final accuracy with only $43\%$ of the training FLOPs (Figure~\ref{fig:gsm8k}). It also outperforms deeper MDMs with comparable per-step training FLOPs, indicating that the gains cannot be explained by depth alone.

\paragraph{Experimental setup.}
LoopMDM is matched to a same-size MDM baseline in parameter count and total training FLOPs, and to the deepest MDM baseline (with 21 layers) in per-step training FLOPs, isolating the contribution of iterative application from raw depth. Following prior work~\citep{kim2026stop}, we use a 125M MDM with Qwen2-style attention~\citep{yang2023qwen2} (hidden size 512, 14 layers, 8 heads, sequence length 512), pretrained on TinyGSM~\citep{liu2023tinygsm}, an 11.8M-problem corpus of GSM8K questions paired with structured Python solutions. LoopMDM loops two mid-layers 1–2 in zero-based indexing with $S_{\max}=8$ and is compared, under matched total training compute, against $14$-, $18$-, and $21$-layer baselines. 
We evaluate accuracy with Top-2 and Top-3 unmasking under maximum token-wise probability, averaging results over three random seeds. Full details are in Appendix~\ref{appendix:expdetail}.

\paragraph{Looping uses depth more effectively than stacking layers.} Figure~\ref{fig:gsm8k} shows that LoopMDM consistently outperforms the $14$-layer baseline under matched training compute, with gains increasing monotonically with the loop count $S$ and consistent across Top-2 and Top-3 unmasking. At $S=16$ (beyond the training maximum $S_{\max}=8$), the improvement reaches $+8.5$ and $+8.1$ points, respectively. LoopMDM also matches the $14$-layer baseline's final accuracy using only $43\%$ of its training FLOPs. More importantly, it outperforms a $21$-layer baseline with comparable per-step FLOPs, achieving gains of $+2.7$ at $S=8$ and $+3.4$ points at $S=16$. These results show that repeated application of a shared block is more effective than increasing depth for the reasoning task.

\subsection{Where do the gains come from?} 
\label{sec:insight}

Sections~\ref{sec:finding-efficiency} and~\ref{sec:finding-math} establish that 
selective looping helps; we now examine \emph{how}. 
He~\etal \citep{he2026reasoning} show that MDMs rely on masked positions whose hidden representations participate in prediction. Now we build on this view to study how looping reshapes the computation over these masked positions. 

First, we present an illustrative separation showing masked positions can expand the class of solvable tasks when used as a workspace. Second, we show that LoopMDM uses masked positions to solve a task that a shallow model cannot, by a controlled Sudoku experiment. Third, we analyze attention patterns and observe increased interactions among masked positions as the loop count grows. Finally, we show these gains vary across timesteps, which motivates a simple adaptive loop allocation strategy.

\paragraph{Masks as workspace: an illustrative separation.}
We present a simple example to illustrate the role of masked positions as a workspace. The goal is not to characterize typical behavior, but to provide a clear separation that makes this intuition concrete. We consider the $k$-Clique problem as a canonical globally coupled task, where all candidate solutions must be evaluated jointly.

\begin{restatable}{Example}{Entropy}
\label{prop}
\textbf{Example.} \emph{There exists a class of globally coupled prediction tasks solvable by a single denoising step with a masked workspace and a constant number of internal loops, but not by an otherwise-identical predictor without masks under the same loop budget.}
\end{restatable}

\emph{Proof sketch.} A LoopMDM can allocate $n^k$ masked positions, one per candidate $k$-tuple, and evaluate all candidates in parallel within a single denoising step. A predictor without masked positions lacks such parallel storage under the same compute budget. The full proof is provided in Appendix~\ref{sec:mdm-theory}.

This example serves to sharpen intuition: masked positions can act as a parallel workspace, enabling forms of computation that are not available without them under the same loop budget.

\begin{figure}
    \centering
    \vspace{-0.5em}
    \includegraphics[width=1\linewidth]{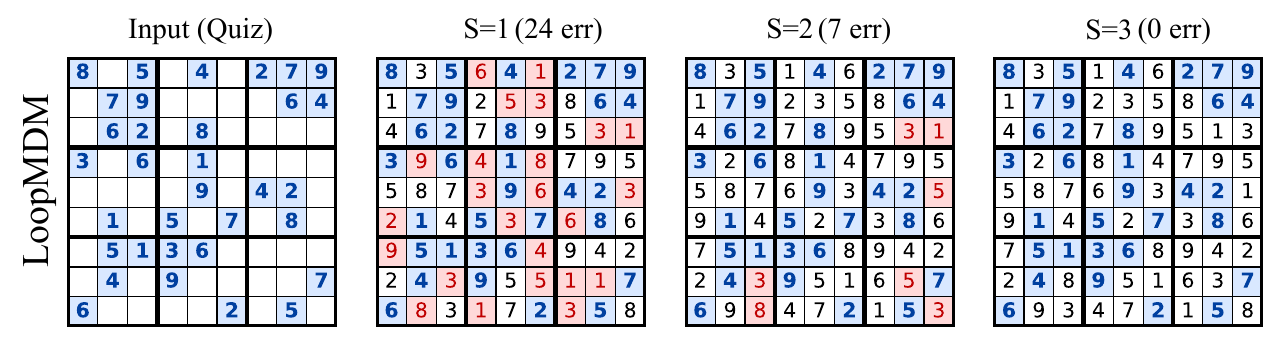}
    \vspace{-1.5em}
    \caption{
\textbf{Looping recovers global consistency under a restricted generation order.}
While MDMs can solve Sudoku using adaptive unmasking, we remove this advantage by enforcing a fixed left-to-right (autoregressive-style) order, where early predictions are made with incomplete context. This isolates the role of within-step computation, since improvements can no longer come from selecting easier cells first. We show predictions at loop counts $S\!=\!1,2,3$ on the same puzzle (red = errors). As $S$ increases, errors sharply decrease ($24\!\to\!7\!\to\!0$), suggesting that looping allows masked positions to revise globally inconsistent early predictions before token commitment.
}
    \label{fig:sudoku_vis}
    \vspace{-1.0em}
\end{figure}

\paragraph{LoopMDM uses masked positions to solve Sudoku under a restricted generation order.} The expressivity result above shows what masks \emph{can} do; we ask whether a trained LoopMDM \emph{does} use them this way in practice. Sudoku provides a natural testbed for this question: solving the puzzle requires assigning a correct value to every position under global consistency constraints, making interactions across unresolved positions essential. MDMs can solve Sudoku near-perfectly when using adaptive unmasking orders that resolve easier cells first~\citep{kim2025train}, but this obscures the role of masked tokens during inference.

To isolate this effect, we enforce a fixed left-to-right generation order, identical to that used in autoregressive models, regardless of cell difficulty. This setting is known to be challenging even for large autoregressive models, including those at the 7B scale~\citep{ye2024beyond}. Under this restriction, any improvement must arise from computation within each denoising step rather than from the generation order. We jointly train a 1-layer MDM baseline and its looped counterpart under the same architecture ($d_{\text{model}}=384$, 12 heads) on a 100k Sudoku dataset, and verify convergence within 100 epochs using $S_{\max}=6$. We then vary the inference-time loop count $S$ and measure the exact-solve rate on a 1,000-test set.

Figure~\ref{fig:sudoku_vis} shows a representative trajectory, where errors decrease from $24$ to $7$ to $0$, as the loop count increases. Early iterations produce locally plausible but globally inconsistent assignments, while later iterations revise these through interactions among masked positions, yielding a consistent solution. 
Meanwhile, the 1-layer MDM baseline achieves only $10.9\%$ accuracy, and the looped model at $S=1$ similarly reaches $10.6\%$, indicating that shallow left-to-right generation alone is insufficient for solving the task.

\paragraph{Mask-to-mask attention increases with loop count in language modeling.}

The Sudoku result shows that LoopMDM can effectively utilize masked positions; we next examine whether similar patterns arise in language modeling. We analyze how interactions among masked positions evolve with the loop count. On a LoopMDM trained on OpenWebText, we measure \emph{mask-to-mask attention} at two mid-block layers, defined as the average attention mass from masked-position queries to masked-position keys, while sweeping the inference loop count $S$ up to and beyond $S_{\max}$.

\begin{figure}
    \centering
    \includegraphics[width=0.9\textwidth]{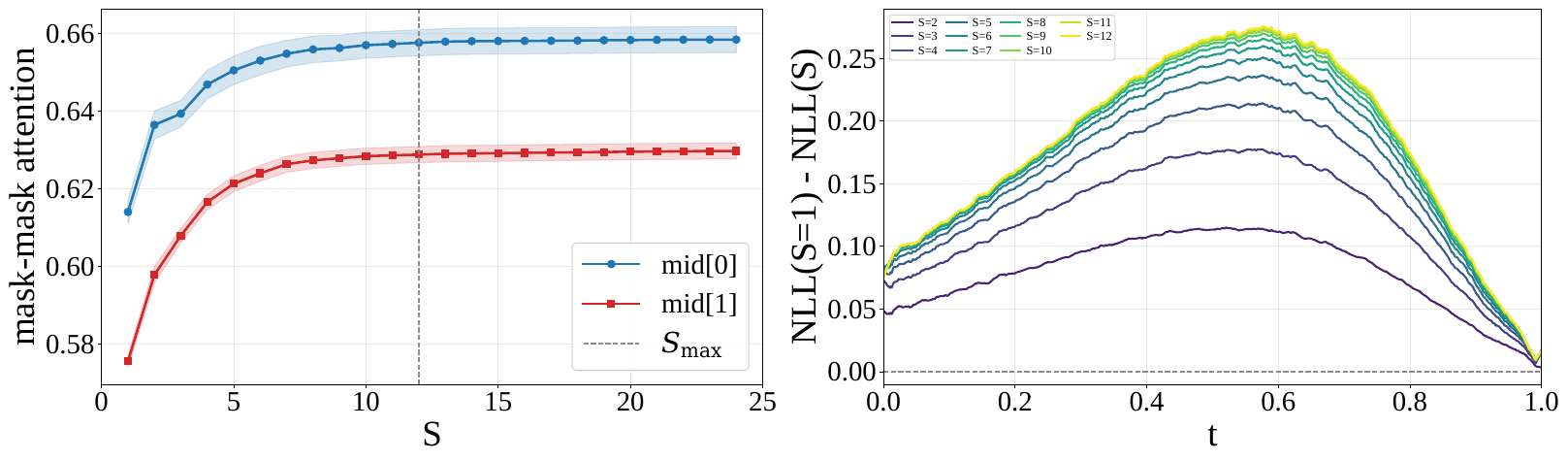}
    \vspace{-0.3cm}
    \caption{
\textbf{Analysis of looping behavior.} 
(Left) Average mask-to-mask attention at timestep $t=0.5$ for a model with $n_m=2$, measured at the two mid-block layers (denoted $\mathrm{mid[0]}$ and $\mathrm{mid[1]}$) as a function of loop counts $S$. Each curve shows attention at the corresponding layer after $S$ loop applications. Attention increases with $S$ and saturates near the training maximum $S_{\max}=12$ (dashed line). 
(Right) NLL improvement $\mathrm{NLL}(S\!=\!1)-\mathrm{NLL}(S)$ across timesteps for sequences of length 1024 on OpenWebText, with $S_{\max}=12$. As $S$ increases, improvement becomes more concentrated and converges faster, indicating that additional loop counts primarily refine intermediate timesteps.
}
    \label{fig:analysis}
    \vspace{-0.5em}
\end{figure}

Figure~\ref{fig:analysis}~(left) shows that mask-to-mask attention increases with $S$ at both layers and plateaus near the training maximum $S_{\max}=12$. The trend is consistent across layers and remains stable when $S$ exceeds $S_{\max}$. While not causal, this pattern suggests that later loop counts place greater emphasis on interactions among masked positions, consistent with the behavior observed in the Sudoku experiment.

\paragraph{Looping helps unevenly across denoising timesteps.}
\label{adaptivelooping}
We next ask where along the denoising process looping is most useful. For each timestep $t$, we measure the per-position NLL gain from additional loop iterations, $\mathrm{NLL}_t(S=1)-\mathrm{NLL}_t(S)$. Figure~\ref{fig:analysis} (right) shows that the gains are not uniform: they peak at intermediate timesteps, where the model has both sufficient context and unresolved positions, and are smaller near the two ends of the denoising process. Increasing $S$ amplifies this profile but gradually saturates, suggesting that additional loops are useful unevenly across timesteps.

\begin{wraptable}{r}{0.46\textwidth}
    \centering
    \vspace{-2.0em}
    \caption{\textbf{Adaptive looping trade-off.}
    Adaptive looping preserves performance while substantially reducing the average loop count.}
    \small
    \resizebox{0.46\textwidth}{!}{%
    \begin{tabular}{lcc}
    \toprule
    Setting & $S \rightarrow \bar{S}$ & Metric \\
    \midrule
    Zero-shot PPL ($\downarrow$) & $12 \rightarrow 5.1$ & $51.7 \rightarrow 52.9$ \\
    Generative PPL ($\downarrow$) & $12 \rightarrow 5.2$ & $38.53 \rightarrow 39.11$ \\
    Downstream Acc. ($\uparrow$) & $12 \rightarrow 4.9$ & $41.4 \rightarrow 41.3$ \\
    \bottomrule
    \end{tabular}}
    \label{tab:adaptive_summary}
    \vspace{-1.0em}
\end{wraptable}

This observation suggests a simple way to choose loop budgets at inference time. Rather than fixing the same loop count at every timestep, we monitor the relative hidden-state change between successive loop iterations, $\|H^{(k)} - H^{(k-1)}\| / \|H^{(k)}\|$, and stop once this ratio falls below a threshold $\epsilon$ or the maximum loop budget is reached. In all experiments, we use $\epsilon=0.1$. This rule allocates fewer iterations when hidden states stabilize early and more when updates remain large. As summarized in Table~\ref{tab:adaptive_summary}, adaptive looping reduces the average loop count from $S=12$ to roughly five iterations while preserving most of the performance gains across all evaluation settings.

Appendix~\ref{add:adaptive} shows that this rule yields a smooth computation-performance trade-off and naturally allocates more loops to intermediate timesteps. These results provide a simple practical guide for selecting loop counts in LoopMDM.

\section{Conclusion}

We introduced LoopMDM, which selectively loops a small shared block inside masked diffusion language models. LoopMDM improves matched-compute performance across language modeling and reasoning tasks without increasing parameter count, and achieves comparable or better performance with fewer training FLOPs than non-looped baselines. These results suggest that the gains from looping are not solely explained by increased depth, but are consistent with benefits from reusing computation within the diffusion architecture. More broadly, our findings indicate that selective looping can serve as a simple approach for improving masked diffusion models under fixed parameter and compute budgets. We believe our work opens up broad opportunities for architectural advances in improving training efficiency and generalization. We provide limitations and future work in Appendix~\ref{limitation}.

\bibliographystyle{plain}
\bibliography{main}
\newpage
\appendix
\section{Related Work}
\label{sec:related}

\paragraph{Masked diffusion models.}
Discrete diffusion models~\citep{austin2021structured, Hoogeboom2021b} model discrete data distributions via a diffusion process, and have recently emerged as a promising alternative to autoregressive language models. 
Two main types of forward corruption processes are commonly used: uniform corruption, which replaces tokens with uniformly random tokens~\citep{lou2024discrete,sahoo2025diffusionduality}, and masking corruption, which replaces tokens with a special mask token~\citep{sahoo2024simple,shi2025simplified}. 
Among these, masked diffusion models (MDMs) have become the dominant framework for discrete diffusion models due to their simplicity and strong empirical performance.

The standard masking-based formulation~\citep{ou2024RADD,sahoo2024simple} remains the most widely adopted backbone. 
Large-scale models such as LLaDA family~\citep{nie2025llada,zhu2025llada,bie2025llada2}, Seed Diffusion~\citep{song2025seed}, and Dream~\citep{ye2025dream} retain the basic generation scheme, where a fully masked sequence is iteratively denoised into tokens. Empirically, MDMs have become increasingly competitive with autoregressive models, with recent models matching or outperforming comparable AR baselines across multiple benchmarks~\citep{nie2025llada,ye2025dream,gong2026diffucoder,xie2025dreamcoder}.

\paragraph{Looped transformers.}
Looping has long been explored as a way to increase the effective depth of
transformers without increasing the number of parameters. Early examples include
the Universal Transformer~\citep{dehghani2018universal}, which repeatedly applies
a shared transition block and shows strong performance on algorithmic tasks, and
ALBERT~\citep{lan2019albert}, which demonstrates the practical benefits of
cross-layer parameter sharing in BERT-style pretraining. Subsequent work has
studied looped transformers as programmable or iterative computation models
\citep{giannou2023looped,yang2023looped}, and more recent large-scale models
such as Ouro~\citep{zhu2025scaling} and Huginn~\citep{geiping2025scaling} suggest
that recurrent depth can scale to modern language modeling regimes.

Broadly, existing work on looped transformers can be viewed along two related
axes. The first emphasizes iterative refinement: repeated applications of the
same block allow the model to progressively improve an internal representation.
This perspective has been used to improve length generalization on algorithmic
tasks~\citep{fan2024looped}, to implement token-level variants of
chain-of-thought-style computation~\citep{mohtashami2023cotformer}, and to study
the theoretical power of repeated transformer computation
\citep{saunshi2025reasoning}. The second emphasizes parameter efficiency:
weight sharing allows a model to obtain greater effective depth under a fixed
parameter budget. This idea predates modern LLMs and has been explored in
recurrent or shared-parameter sequence models~\citep{dabre2019recurrent,
takase2023lessons}, while recent work shows that looped language models can
achieve strong reasoning ability under iso-FLOP or parameter-efficient regimes
\citep{saunshi2025reasoning,zhu2025scaling, prairie2026parcae}.

Our work combines these two perspectives in the setting of masked diffusion
language models. Rather than using looping only to reduce parameters or to add
generic latent depth, we study how looped computation interacts with the
distinct structure of MDMs: the presence of many masked positions. These masked
positions provide a latent workspace, while the looped block iteratively refines
the hidden states on that workspace before token prediction. This lets us obtain
both training efficiency through selective weight sharing and iterative
refinement through computation over masked positions, which is not naturally present in standard autoregressive looping, where unresolved output positions are not maintained as an explicit masked workspace.

\paragraph{Test-Time Scaling in Masked Diffusion.}
Scaling inference-time computation has emerged as a complementary axis to model scaling. In autoregressive models, chain-of-thought prompting~\citep{wei2022chain} increases computation by generating longer reasoning traces. Subsequent work explores more implicit forms of test-time compute, including pause or filler tokens~\citep{goyal2024think,pfau2024let} and latent-space reasoning that does not rely on explicit natural language tokens~\citep{deng2023implicit,hao2024training}. These results suggest that additional computation can be effectively allocated without being directly exposed in the output sequence.

In masked diffusion, most existing test-time scaling methods focus on the sampling procedure itself. Key design choices include how many tokens to unmask at each step, which positions to select, and whether to revise previously generated tokens. While confidence-based heuristics are commonly used, recent work shows that the unmasking schedule critically affects generation quality. This has led to adaptive timestep or parallel decoding strategies~\citep{israel2025accelerating}, dilated unmasking schedules that avoid resolving strongly coupled tokens simultaneously~\citep{luxembourg2025plan}, remasking-based samplers that enable correction during inference~\citep{wang2025remasking}, and search or planned unmasking policies~\citep{peng2025path,peng2025planner,lee2025lookahead,lee2025effective,bilal2026s}.

\section{Theoretical Analysis}
\label{sec:mdm-theory}
\subsection{Additional Background: Padded Looped Transformers and Masked Diffusion}
\label{app:background-plt-mdm}

We briefly review the theoretical perspective that motivates our interpretation
of masked positions as computational workspace.

\paragraph{Padding and looping as computational resources.}
Recent work on padded looped transformers studies blank or padding tokens as a form of
parallel test-time computation. These padding tokens do not carry additional
input content; rather, they can
store intermediate information. In this sense, padding behaves like a
workspace, or equivalently like an increase in computational width. Looping, on
the other hand, repeatedly applies a shared transformer block and therefore
increases effective computational depth without increasing the number of
parameters. This distinction is explicit in the theory of padded looped
transformers: padding controls the amount of parallel storage available to the
model, while looping controls the number of iterative refinement steps~\citep{merrill2025exact}.

\paragraph{Connection to masked diffusion models.}
This viewpoint has recently been connected to masked diffusion language models
(MDMs). In the finite-precision logarithmic-width setting, MDMs and padded
looped transformers (PLTs) are closely related computational models~\citep{svete2025reasoning}. At a high level, both models perform iterative
parallel refinement. A PLT refines continuous residual states over an input
augmented with padding positions, while an MDM refines a partially masked or
partially generated sequence through denoising steps. Under this correspondence,
the generated or masked positions of an MDM play a role analogous to padding
positions in a PLT: they provide additional locations where intermediate
computation can be represented before the final prediction is read out.

More concretely, prior work shows that MDMs can be simulated by PLTs with
comparable step and output budgets, and that PLTs can be simulated by MDMs with
additional output space used to store the PLT residual stream~\citep{svete2025reasoning}.
Although the two formalisms differ in how information is represented, with PLTs
using continuous residual states and MDMs operating over discrete masked or
generated symbols across denoising steps, they share the same core computational
structure. In both cases, additional positions provide storage for intermediate
information, and repeated application of a model block refines that stored state.

\paragraph{Relation to our setting.}
Our model differs slightly from the standard PLT and MDM correspondence. We place
looping inside a single MDM denoising network rather than increasing the number
of external denoising steps. Nevertheless, the same conceptual decomposition is
useful. The masked positions are present throughout the denoising network's
forward computation, and their hidden states can be updated repeatedly by the
looped mid-block before the model commits to discrete token predictions. We
therefore view the masked positions as a latent workspace, and the internal
loops as iterative computation over this workspace.

We use this perspective only as an explanatory tool. The goal of the separation
below is not to give a tight characterization of practical LoopMDMs, nor to
claim that all assumptions of the formal PLT and MDM models hold exactly in our
architecture. Rather, the goal is to isolate a simple mechanism: masked
positions can supply parallel workspace that a padding-free model lacks.

\subsection{The \(k\)-Clique Workspace Task}
\label{app:kclique-task}
Let \(G\) be an \(n\)-vertex graph represented by its adjacency matrix
\[
A\in\{0,1\}^{n\times n}.
\]
The input length is \(N=\Theta(n^2)\). For a fixed constant \(k\), define
\[
\mathrm{CLIQUE}_{k,n}(A)=1
\]
if and only if there exist distinct vertices
\[
v_1,\dots,v_k\in[n]
\]
such that
\[
A_{v_r,v_s}=1
\qquad
\text{for all }1\le r<s\le k.
\]

In the padded version of the task, we append \(n^k\) masked workspace positions
and one masked answer position. The \(n^k\) workspace positions are indexed by
ordered tuples
\[
\mathbf v=(v_1,\dots,v_k)\in[n]^k.
\]
The intended role of the workspace position indexed by \(\mathbf v\) is to store
whether this tuple forms a \(k\)-clique.

\subsection{Main Example}
\label{app:kclique-example}

\begin{example}
\label{thm:kclique-workspace-separation}
Fix a padding-free finite-precision logarithmic-width transformer architecture
with a constant number of internal loop counts. Suppose that, on inputs of
length \(N\), this architecture is simulable by constant-depth \(\mathrm{AC}^0\)
circuits of size \(O(N^C)\) for some constant \(C\).

Then there exists a constant \(k>8C\) such that the corresponding padding-free
architecture with no additional positions does not decide
\(\mathrm{CLIQUE}_{k,n}\) on \(n\)-vertex graphs.

On the other hand, there exists a one-step masked-workspace transformer
architecture with \(n^k+O(1)\) masked positions, the same asymptotic
finite-precision and logarithmic-width setting, a constant loop budget, and
constant local resources depending only on \(k\), that decides
\(\mathrm{CLIQUE}_{k,n}\).
\end{example}

\begin{proof}[Proof of Example]

The key distinction is that masked workspace provides an explicit set of
$n^k$ hypothesis slots, one for each candidate $k$-tuple of vertices.
This allows the model to evaluate all candidates in parallel.

In contrast, a base transformer must encode all intermediate
combinatorial structure within a logarithmic-width representation.
Under finite precision and a constant loop budget, this restricts the
model to computations simulable by constant-depth $\mathrm{AC}^0$ circuits,
which are known to be insufficient for solving $k$-Clique for large enough $k$.\\

\textbf{Upper bound: masked workspace solves the task.}
\label{app:kclique-upper}
We construct a padded looped transformer that decides \(\mathrm{CLIQUE}_{k,n}\).

The graph is represented by edge tokens \(e_{a,b}\), one for each ordered pair
\((a,b)\in[n]^2\). The token \(e_{a,b}\) stores the edge bit \(A_{a,b}\), and
its positional encoding stores the pair \((a,b)\).

We append \(n^k\) masked workspace positions and one masked answer position.
Each workspace position is indexed by a tuple
\[
\mathbf v=(v_1,\dots,v_k)\in[n]^k.
\]
For example, the \(t\)-th workspace position can be assigned the tuple whose
coordinates are
\[
v_r(t)
=
1+
\left\lfloor
\frac{t-1}{n^{r-1}}
\right\rfloor
\bmod n,
\qquad r=1,\dots,k.
\]
Since \(k\) is constant, this tuple information requires only \(O(k\log n)\)
bits, which is compatible with logarithmic width.

For every pair \(1\le r<s\le k\), the workspace token \(z_{\mathbf v}\)
uses one attention head to retrieve the edge token \(e_{v_r,v_s}\). This can be
implemented by position-matching attention: the query at \(z_{\mathbf v}\)
encodes the pair \((v_r,v_s)\), while the key at edge token \(e_{a,b}\) encodes
\((a,b)\). Hence the head retrieves the value \(A_{v_r,v_s}\).

After retrieving all edge bits
\[
A_{v_r,v_s},
\qquad
1\le r<s\le k,
\]
a position-wise feedforward network computes
\[
c_{\mathbf v}
=
\left(
\bigwedge_{1\le r<s\le k} A_{v_r,v_s}
\right)
\wedge
\left(
\bigwedge_{1\le r<s\le k} [v_r\ne v_s]
\right).
\]
Because \(k\) is constant, this is a constant-size Boolean computation. Thus
\(c_{\mathbf v}=1\) if and only if the tuple \(\mathbf v\) forms a valid
\(k\)-clique.

The answer position then attends to all workspace positions and computes
\[
\bigvee_{\mathbf v\in[n]^k} c_{\mathbf v}.
\]
This value is \(1\) exactly when the graph contains a \(k\)-clique.

\textbf{Lower bound: base constant-loop models cannot solve the task.}
\label{app:kclique-lower}
We prove the lower bound for the constant \(k>8C\) chosen in
Example~\ref{thm:kclique-workspace-separation}. Suppose, for contradiction, that
the corresponding base looped transformer decides
\(\mathrm{CLIQUE}_{k,n}\).

The padding-free model receives only the graph tokens. Since the graph is
represented by an \(n\times n\) adjacency matrix, its input length is
\[
N=\Theta(n^2).
\]
Fixed padding-free
architecture is simulable, on inputs of length \(N\), by a constant-depth
\(\mathrm{AC}^0\) circuit of size
\[
O(N^C).
\]
The constant loop budget is already accounted for in this hypothesis: unrolling
a constant number of internal loop counts only changes the effective depth
by a constant depending on the fixed architecture.

Therefore, if the padding-free transformer decided
\(\mathrm{CLIQUE}_{k,n}\), then there would exist a constant-depth
\(\mathrm{AC}^0\) circuit family deciding \(\mathrm{CLIQUE}_{k,n}\) with size
\[
O(N^C)=O\bigl((n^2)^C\bigr)=O(n^{2C}).
\]

We now relate this ordered-adjacency representation to the standard undirected
\(k\)-Clique problem. Rossman's lower bound~\citep{rossman2008constant} is stated for simple undirected
graphs. If a circuit decides \(\mathrm{CLIQUE}_{k,n}\) on ordered adjacency
matrices \(A\in\{0,1\}^{n\times n}\), then it also yields a circuit for the
standard undirected representation with variables
\(\{x_{\{u,v\}}:1\le u<v\le n\}\) by substituting
\[
A_{u,v}=A_{v,u}=x_{\{u,v\}}
\qquad\text{for }u<v,
\]
and
\[
A_{u,u}=0
\qquad\text{for all }u\in[n].
\]
This substitution only hard-wires and duplicates input variables; it does not
increase the circuit depth or asymptotic size. Hence the assumed padding-free
transformer would imply a constant-depth \(\mathrm{AC}^0\) circuit family of
size \(O(n^{2C})\) for the standard \(k\)-Clique problem on \(n\)-vertex
graphs.

Rossman's lower bound states~\citep{rossman2008constant} that, for every fixed constant \(k\), the
\(k\)-Clique problem on \(n\)-vertex graphs requires constant-depth circuits of
size
\[
\omega(n^{k/4}).
\]
Choose \(k>8C\). Then
\[
\frac{k}{4}>2C,
\]
and therefore
\[
O(n^{2C}) = o(n^{k/4}).
\]
This contradicts Rossman's lower bound. Thus the padding-free constant-loop
predictor cannot decide \(\mathrm{CLIQUE}_{k,n}\).
\end{proof}

\section{Experimental Details}
\label{appendix:expdetail}
\subsection{Experimental details for language modeling}
\paragraph{Datasets and preprocessing.}
We evaluate on One Billion Word (LM1B)~\citep{chelba2013lm1b}, OpenWebText (OWT)~\citep{Gokaslan2019OpenWeb}, and FineWeb-Edu~\citep{penedo2024fineweb}. 
For LM1B, we follow prior work~\citep{lou2024discrete, sahoo2024simple} and detokenize the dataset, then tokenize using the \textsc{bert-base-uncased} tokenizer~\citep{he2022diffusionbert}. 
Sequences are concatenated and packed to a fixed length of 128 using sentence packing~\citep{raffel2020exploring}. 
For OWT and FineWeb-Edu, we use the GPT-2 tokenizer~\citep{radford2019language} and pack sequences to a context length of 1{,}024, inserting an \texttt{eos} token between documents. 
Following prior work, we reserve the last 100K documents of OWT for validation; FineWeb-Edu uses its standard split.

\paragraph{Model and training.}
We use a 170M-parameter diffusion transformer (DiT)~\citep{peebles2023scalable} with rotary positional encoding and adaptive layer normalization, consistent across all datasets. 
The model has 12 layers with hidden size 768 and 12 attention heads. 
For LoopMDM, we adopt an iso-parameter setting by looping two layers 1–2 in zero-based indexing (starting from the second layer) with a maximum loop count $S_{\max}=12$. All experiments were conducted using 4 NVIDIA A100 GPUs.

The baseline is trained for 1M steps; LoopMDM is trained for proportionally fewer steps to match total training FLOPs with batch size 512 using the AdamW optimizer. 
We use a linear warmup schedule to $3\mathrm{e}{-4}$ over 2,500 steps, following prior works~\citep{sahoo2024simple,lou2024discrete,sahoo2025diffusionduality}.  All looped models are trained under matched training FLOPs to ensure fair comparison with the baseline.

\subsection{Experimental details of math reasoning}
\paragraph{Dataset.}
We use TinyGSM~\citep{liu2023tinygsm} as the pretraining corpus, which consists of 11.8M GSM8K-style problems paired with structured Python solutions. 
Following prior works~\citep{liu2023tinygsm,kim2026stop}, this formulation improves learnability at small model scales and enables controlled comparison of reasoning performance.

\paragraph{Model and training.}
We use a 125M-parameter masked diffusion model with Qwen2-style attention, consisting of 14 layers, hidden size 512, 8 attention heads, and maximum sequence length 512. 
All models share the same architecture, optimizer, and training setup. All experiments were conducted using NVIDIA A100 GPUs.

For LoopMDM, we adopt an iso-parameter setting by looping two layers 1–2 in zero-based indexing, while all other components remain unchanged. 
We train for 900k iterations using AdamW with learning rate $3\mathrm{e}{-4}$, batch size 32 per GPU on 4 GPUs, and apply exponential moving average (EMA) with decay 0.9999. 

\paragraph{Evaluation.}
Evaluation is performed on the GSM8K test set every 5k steps using Top-2 and Top-3 unmasking with maximum token-wise probability as confidence. All experiments were conducted using NVIDIA A100 GPUs.
All results are reported using EMA checkpoints. 

\subsection{Experimental details of downstream task evaluation}

We evaluate downstream performance using the models pretrained on FineWeb-Edu for 1M steps under the same setup as described in the language modeling experiments. 
No additional fine-tuning is performed.

\paragraph{Benchmarks.}
We evaluate on nine widely used benchmarks spanning three categories. 
For \textit{commonsense reasoning}, we use HellaSwag~\citep{zellers2019hellaswag}, 
PIQA~\citep{bisk2020piqa}, 
SIQA~\citep{sap2019social}, 
and WinoGrande~\citep{sakaguchi2021winogrande}. 
For \textit{reading comprehension}, we use 
BoolQ~\citep{clark2019boolq} and RACE~\citep{lai2017race}. 
For \textit{scientific question answering}, we use 
OpenBookQA~\citep{mihaylov2018can}, 
ARC-Easy, and ARC-Challenge~\citep{clark2018think}.

\paragraph{Evaluation protocol.}
We adopt a likelihood-based evaluation. 
For each multiple-choice question, we estimate the likelihood of each candidate answer by conditioning on the input context and selecting the option with the highest likelihood. 
Specifically, we compute likelihoods over up to 128 tokens per candidate. 
Since the model is trained with a maximum sequence length of 1024, all inputs are padded to this length during evaluation.

\subsection{Experimental details of analysis}

\paragraph{Mask count analysis.}
To analyze the effect of looping across different masking regimes, we partition sequences by the number of masked positions. 
Specifically, we vary the number of masked tokens from 1 to 1024 and group sequences accordingly. 
For each mask count, we measure the improvement in NLL on the OpenWebText validation set, comparing different inference-time loop counts.

\paragraph{Mask-to-mask attention analysis.}
To study how looping affects interactions between masked positions, we analyze attention patterns across loop counts. 
We sample 10,000 timesteps uniformly from $t \in [0,1]$ and compute the average attention weights at each loop counts. 
In particular, we measure mask-to-mask attention by averaging the attention mass assigned from masked-query tokens to masked-key tokens.

\section{Omitted Results}

\begin{table}[t]
\centering
\caption{\textbf{Effect of the adaptive stopping threshold $\epsilon$.}
Smaller $\epsilon$ increases the average number of loop counts and improves performance, while larger $\epsilon$ reduces compute at the cost of degraded perplexity. 
We use $\epsilon=0.1$ in the main paper as a balanced operating point, retaining most of the gains of fixed $S=12$ while reducing average compute by roughly half.}
\label{tab:epsilon}
\small
\setlength{\tabcolsep}{4pt}
\renewcommand{\arraystretch}{1.05}

\begin{tabular}{cc}
\begin{minipage}{0.48\linewidth}
\centering
\begin{tabular}{lccc}
\toprule
\multicolumn{4}{c}{\textbf{Zero-shot PPL Ablation}} \\
\midrule
Mode & Avg PPL $\downarrow$ & $\bar{S}$ $\downarrow$ & Compute \\
\midrule
$S=6$ & 56.80 & 6.00 & 50.0\% \\
 $S=12$ & 55.58 & 12.00 & 100.0\% \\
$\epsilon=0.02$ & 55.57 & 11.73 & 97.7\% \\
$\epsilon=0.04$ & 55.68 & 9.60 & 80.0\% \\
$\epsilon=0.06$ & 55.94 & 8.02 & 66.9\% \\
$\epsilon=0.08$ & 56.28 & 6.96 & 58.0\% \\
$\epsilon=0.10$ & 56.68 & 6.20 & 51.6\% \\
$\epsilon=0.12$ & 57.16 & 5.49 & 45.7\% \\
$\epsilon=0.14$ & 57.64 & 5.02 & 41.8\% \\
$\epsilon=0.16$ & 58.21 & 4.47 & 37.3\% \\
$\epsilon=0.18$ & 58.81 & 4.12 & 34.3\% \\
$\epsilon=0.20$ & 59.45 & 3.74 & 31.2\% \\

\bottomrule
\end{tabular}
\end{minipage}
&
\begin{minipage}{0.48\linewidth}
\centering
\begin{tabular}{lccc}
\toprule
\multicolumn{4}{c}{\textbf{Generative PPL Ablation}} \\
\midrule
Mode & Gen PPL $\downarrow$ & $\bar{S}$ $\downarrow$ & Compute \\
\midrule
 $S=6$ & 63.60 & 6.00 & 50.0\% \\
$S=12$ & 62.44 & 12.00 & 100.0\% \\
$\epsilon=0.02$ & 62.66 & 10.73 & 89.5\% \\
$\epsilon=0.04$ & 64.17 & 8.33 & 69.4\% \\
$\epsilon=0.06$ & 63.16 & 7.10 & 59.2\% \\
$\epsilon=0.08$ & 64.11 & 6.21 & 51.7\% \\
$\epsilon=0.10$ & 64.39 & 5.45 & 45.4\% \\
$\epsilon=0.12$ & 65.04 & 4.86 & 40.5\% \\
$\epsilon=0.14$ & 64.68 & 4.52 & 37.7\% \\
$\epsilon=0.16$ & 64.93 & 3.94 & 32.9\% \\
$\epsilon=0.18$ & 65.51 & 3.74 & 31.2\% \\
$\epsilon=0.20$ & 66.03 & 3.41 & 28.4\% \\
\bottomrule
\end{tabular}
\end{minipage}
\end{tabular}
\end{table}
\subsection{Ablation on Adaptive Looping}
\label{add:adaptive}

\paragraph{Trade-off controlled by the threshold $\epsilon$.}
Table~\ref{tab:epsilon} reports the effect of the adaptive stopping threshold $\epsilon$ on average zero-shot perplexity and generative perplexity. Smaller $\epsilon$ increases the average number of loop counts and improves performance, while larger $\epsilon$ reduces computation with gradually degraded perplexity. Across a broad range of thresholds, the adaptive strategy remains competitive with fixed high loop counts while using substantially fewer iterations on average.

In the main paper, we use $\epsilon=0.1$, which provides a balanced trade-off between performance and computation. At this operating point, adaptive looping reduces the average number of iterations to roughly half of that for a fixed $S=12$ while maintaining comparable zero-shot and generative perplexity.

\begin{figure}[t]
    \centering
    \includegraphics[width=\linewidth]{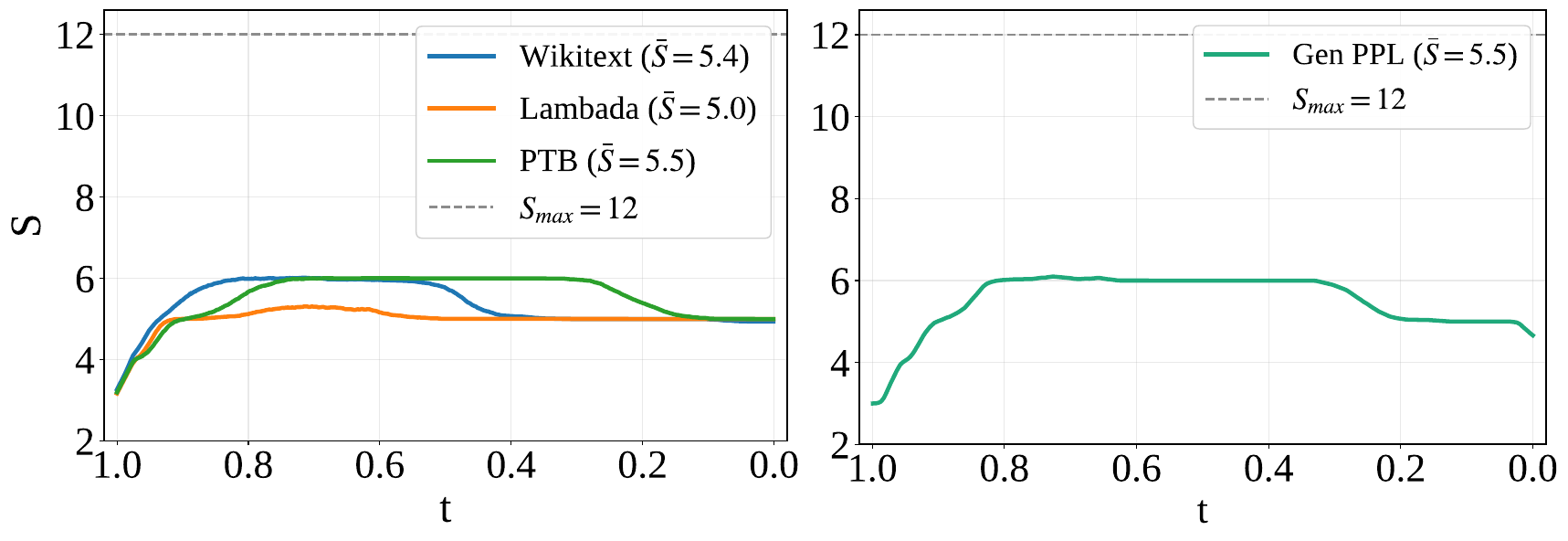}
    \caption{
    \textbf{Adaptive allocation of loop counts across timesteps.}
    Average loop count allocated by the adaptive strategy as a function of timestep $t$ on OpenWebText with $\epsilon=0.1$, measured over 100 sampled sequences. 
    (Left) Zero-shot perplexity evaluation on WikiText, Lambada, and PTB. 
    (Right) Generative perplexity evaluation. 
    Across both settings, the adaptive strategy allocates more iterations at intermediate timesteps and fewer at early and late timesteps. 
    The dashed line indicates the maximum training loop count $S_{\max}=12$.
    }
    \label{fig:adaptive_allocation}
\end{figure}

\paragraph{Adaptive allocation across timesteps.}
Figure~\ref{fig:adaptive_allocation} shows the average number of loop counts allocated by the adaptive strategy as a function of timestep. We evaluate the allocation behavior on OpenWebText using $\epsilon=0.1$, averaging over 100 sampled sequences for both zero-shot perplexity and 1024-step generative perplexity settings.

Across all evaluations, the adaptive strategy consistently allocates more iterations to intermediate timesteps and fewer to early and late timesteps. This matches the analysis in Section~4.3, where looping is most effective when both context and unresolved positions are sufficiently available. The average loop count remains substantially below the training maximum $S_{\max}=12$, while still concentrating computation on the timesteps where looping provides the largest gains.

Together, these results suggest that adaptive looping provides an efficient and interpretable compute allocation strategy, automatically redistributing computation toward the most useful stages of the denoising process.

\subsection{Compare With Other Looping Methods}
\label{app:compare}
\paragraph{Compare with input injection.}
In ARMs, particularly recurrent architectures inspired by equilibrium models~\citep{anil2022path}, input injection is commonly used to stabilize iterative computation~\citep{geiping2025scaling}. In MDMs, predicted positions already begin from a shared absorbing state through the mask token. A natural analogue in this setting is therefore to inject Gaussian noise into the mask embedding during training.

\begin{wrapfigure}{r}{0.38\textwidth}
    \centering
    \vspace{-0.5em}
    \includegraphics[width=\linewidth]{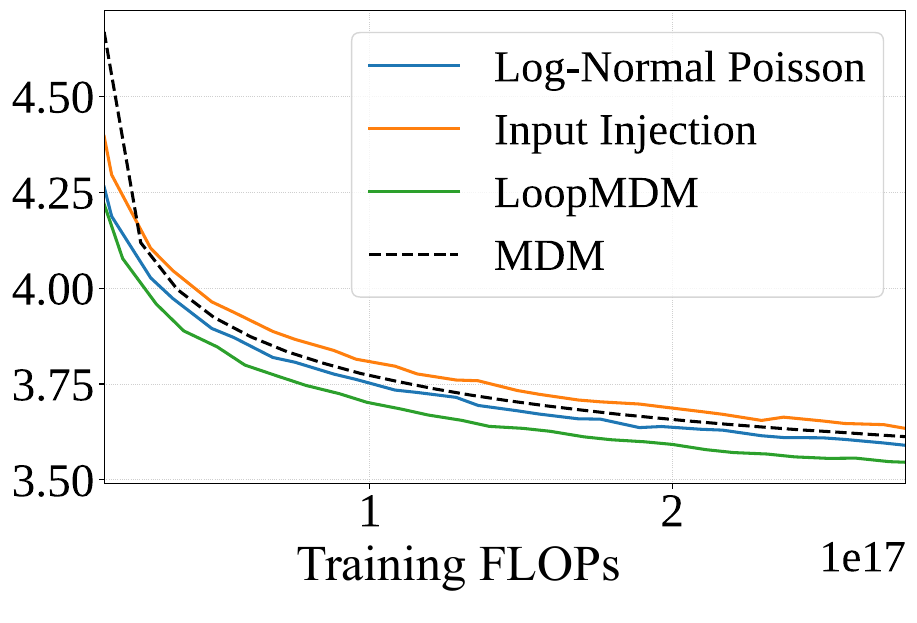}
    \vspace{-1.5em}
    \caption{\small \textbf{NLL comparison with alternative looping strategies.}}
    \label{fig:compare_loop_methods}
    \vspace{-2.0em}
\end{wrapfigure}

As shown in Figure~\ref{fig:compare_loop_methods}, input injection degrades performance relative to the standard MDM baseline. This suggests that techniques developed for recurrent AR models do not transfer directly to the masked diffusion setting.

\paragraph{Compare with log-normal Poisson loop sampling.}
Recent looped transformers sample loop counts from a log-normal Poisson distribution to stabilize recurrent computation and improve test-time scaling~\citep{geiping2025scaling,mcleish2025teaching}. We compare this strategy against the uniform loop-count sampling used in LoopMDM.

Figure~\ref{fig:compare_loop_methods} shows that log-normal Poisson sampling yields only marginal improvement over the baseline and consistently underperforms the uniform sampling strategy used in LoopMDM. These results suggest that the masked diffusion setting benefits more from controlled exposure to different effective depths than from additional stochasticity in loop allocation.

\subsection{Qualitative examples} 
\label{appendix:qual} 
To better understand how looping changes token predictions across iterations, we analyze per-token trajectories as the inference-time loop count increases. Figure~\ref{fig:loop-qualitative} tracks both the predicted token and its NLL at a focal position for several representative examples from OpenWebText. 

Across successful cases (A--D), predictions progressively evolve toward the ground-truth token as the number of loop counts increases. Different behaviors emerge across examples. In some cases, looping reinforces local copying patterns (A), while in others it resolves semantic ambiguity (B), refines syntactic category selection (C), or integrates broader contextual information (D). In all cases, lower NLL correlates with increased confidence in the final prediction.

At the same time, looping is not uniformly beneficial (E) shows a counterexample where additional iterations eventually move the prediction away from an initially correct token. This suggests that repeated application of the shared block can occasionally oversmooth or destabilize local predictions at large loop counts.

\begin{figure*}[t]
\small\centering
\setlength{\tabcolsep}{4pt}
\renewcommand{\arraystretch}{1.05}

\noindent\makebox[\linewidth][c]{%
\fbox{%
\begin{minipage}{\dimexpr.85\linewidth-2\fboxsep-2\fboxrule\relax}
\textbf{(A) Rescued — local copying.}
The focal slot lies inside
\textit{``half \blank{} school \blank{} \textbf{[focal]} book club''};
the loop ends up copying the repeated word.

\par\smallskip
\noindent\textbf{GT:} \texttt{half}

{\footnotesize
\begin{tabular}{l|ccccccc}
$S$  & 1 & 2 & 3 & 4 & \textbf{6} & 12 & 24 \\\hline
pred & a & a & a & a & \textbf{half}\,\checkmark
     & half\,\checkmark & half\,\checkmark \\
NLL  & 7.00 & 4.44 & 2.62 & 1.25 & \textbf{0.38} & 0.13 & 0.13
\end{tabular}
}
\end{minipage}%
}%
}\\[6pt]

\noindent\makebox[\linewidth][c]{%
\fbox{%
\begin{minipage}{\dimexpr.85\linewidth-2\fboxsep-2\fboxrule\relax}
\textbf{(B) Rescued — semantic selection.}
\textit{``\textellipsis at the numbers \blank{} \textbf{[focal]} delegates
are \blank{} proportionately, it's \blank{} how much \blank{} change\textellipsis''}
The loop replaces \texttt{which} with the causal conjunction.

\par\smallskip
\noindent\textbf{GT:} \texttt{Because}

{\footnotesize
\begin{tabular}{l|cccccc}
$S$  & 1 & 2 & \textbf{3} & 4 & 12 & 24 \\\hline
pred & which & which & \textbf{Because}\,\checkmark
     & Because\,\checkmark & Because\,\checkmark
     & Because\,\checkmark \\
NLL  & 6.44 & 4.00 & \textbf{1.75} & 1.12 & 0.50 & 0.38
\end{tabular}
}
\end{minipage}%
}%
}\\[6pt]

\noindent\makebox[\linewidth][c]{%
\fbox{%
\begin{minipage}{\dimexpr.85\linewidth-2\fboxsep-2\fboxrule\relax}
\textbf{(C) Rescued — syntactic refinement (POS).}
\textit{``\textellipsis Geers said. \blank{} \textbf{[focal]} present moment\textellipsis''}
Early iterations emit prepositions; later it locks onto the verb.

\par\smallskip
\noindent\textbf{GT:} \texttt{described}

{\footnotesize
\begin{tabular}{l|ccccccc}
$S$  & 1 & 2 & 3 & 4 & 6 & \textbf{8} & 24 \\\hline
pred & at & at & in & in & in
     & \textbf{described}\,\checkmark
     & described\,\checkmark \\
NLL  & 9.38 & 8.06 & 6.12 & 4.00 & 2.62 & \textbf{1.88} & 1.38
\end{tabular}
}
\end{minipage}%
}%
}\\[6pt]

\noindent\makebox[\linewidth][c]{%
\fbox{%
\begin{minipage}{\dimexpr.85\linewidth-2\fboxsep-2\fboxrule\relax}
\textbf{(D) Rescued — multi-hop / global context.}
\textit{``\textellipsis I had \blank{} Clinton last \blank{} Michigan and saw
\textbf{[focal]} \blank{} a chance to \blank{} off \blank{}-from-behind\textellipsis''}
The model resolves to a topic-specific weekday.

\par\smallskip
\noindent\textbf{GT:} \texttt{Tuesday}

{\footnotesize
\begin{tabular}{l|ccccccc}
$S$  & 1 & 2 & 4 & 6 & 8 & \textbf{12} & 24 \\\hline
pred & her & the & the & the & the
     & \textbf{Tuesday}\,\checkmark
     & Tuesday\,\checkmark \\
NLL  & 7.56 & 5.12 & 2.88 & 2.00 & 1.12 & \textbf{0.62} & 0.38
\end{tabular}
}
\end{minipage}%
}%
}\\[6pt]

\noindent\makebox[\linewidth][c]{%
\fbox{%
\begin{minipage}{\dimexpr.85\linewidth-2\fboxsep-2\fboxrule\relax}
\textbf{(E) Lost — counter-example.}
\textit{``\textellipsis voters were split on candidates' key \blank{}
\textbf{[focal]} Clinton \blank{} as \blank{} electable\textellipsis''}
Further looping drifts to \texttt{whether}.

\par\smallskip
\noindent\textbf{GT:} \texttt{with}

{\footnotesize
\begin{tabular}{l|ccccccc}
$S$  & \textbf{1} & 2 & 3 & 4 & 6 & 8 & 24 \\\hline
pred & \textbf{with}\,\checkmark & with\,\checkmark
     & with\,\checkmark & with\,\checkmark
     & who & whether & whether \\
NLL  & \textbf{0.75} & 0.50 & 0.63 & 0.75 & 2.38 & 5.31 & 6.69
\end{tabular}
}
\end{minipage}%
}%
}

\caption{\textbf{Per-token loop trajectories across iterations.}
Each panel tracks the predicted token (pred) and its NLL at the focal position as the number of loop counts $S$ increases.
Lower NLL indicates higher model confidence.
(A–D) show successful refinement where predictions evolve toward the ground-truth token (GT) via different mechanisms: (A) local copying, (B) semantic selection, (C) syntactic refinement, and (D) global context integration.
(E) illustrates a failure case where further iterations cause drift away from a correct early prediction despite initially low NLL.}
\label{fig:loop-qualitative}
\end{figure*}

\section{Limitations and Future Work}
\label{limitation}

Our work focuses on a simple form of selective looping, where the looped block and adaptive stopping rule are manually specified. While this design already yields consistent improvements across multiple settings, more flexible strategies for allocating computation remain unexplored. In particular, learning where and how much to loop as a function of timestep, token, or input complexity may further improve the efficiency of masked diffusion models.

In addition, our experiments primarily study medium-scale diffusion language models. Recent progress in large-scale MDMs suggests that scaling behavior can change substantially with model size, raising important questions about how selective looping interacts with larger models, longer contexts, and more advanced diffusion architectures. Extending LoopMDM to billion-scale diffusion language models remains an important direction for future work.

Our analysis focuses on attention patterns and controlled reasoning tasks as evidence for how looping interacts with masked positions. Developing more precise theoretical or mechanistic characterizations of these effects, particularly regarding why selective looping in early-middle layers is effective, remains an open problem.

Finally, our work studies looping largely in isolation from existing test-time scaling and search methods for diffusion language models. Exploring how selective looping interacts with techniques such as adaptive unmasking~\citep{yang2026improving, lee2025lookahead}, remasking~\citep{wang2025remasking, zhai2026core}, or search-based decoding~\citep{dang2025inference, lee2025effective} may lead to more effective combinations of architectural and inference-time computation.

\section{Dataset Licenses and Usage Terms}
\label{license}
Table~\ref{tab:asset_licenses} summarizes the datasets and benchmarks used in this work, together with their corresponding licenses and usage terms.
\begin{table}[t]
\centering
\small
\caption{Datasets and benchmarks used in this work, together with their licenses and usage terms.}
\label{tab:asset_licenses}

\begin{tabular}{p{0.28\linewidth} p{0.22\linewidth} p{0.42\linewidth}}
\toprule
\textbf{Dataset / Benchmark} & \textbf{Usage} & \textbf{License / Usage Terms} \\
\midrule

\multicolumn{3}{l}{\textit{Pretraining}} \\
LM1B~\citep{chelba2013lm1b}
& Pretraining
& Not explicitly specified on the official benchmark page; derived from WMT 2011 News Crawl data. \\

OpenWebText~\citep{Gokaslan2019OpenWeb}
& Pretraining
& CC0-1.0 for dataset packaging. \\

FineWeb-Edu~\citep{penedo2024fineweb}
& Pretraining
& ODC-By 1.0; CommonCrawl Terms of Use apply. \\

\midrule

\multicolumn{3}{l}{\textit{Downstream evaluation}} \\

RACE~\citep{lai2017race}
& Downstream evaluation
& Non-commercial research use only. \\

WinoGrande~\citep{sakaguchi2021winogrande}
& Downstream evaluation
& CC-BY for dataset (version unspecified); Apache-2.0 for code. \\

HellaSwag~\citep{zellers2019hellaswag}
& Downstream evaluation
& MIT. \\

Social IQA~\citep{sap2019social}
& Downstream evaluation
& CC-BY-4.0. \\

PIQA~\citep{bisk2020piqa}
& Downstream evaluation
& Academic Free License v3.0 (AFL-3.0). \\

BoolQ~\citep{clark2019boolq}
& Downstream evaluation
& CC BY-SA 3.0. \\

OpenBookQA~\citep{mihaylov2018can}
& Downstream evaluation
& Apache-2.0 for repository; dataset license not separately specified. \\

ARC-Easy /-Challenge~\citep{clark2018think}
& Downstream evaluation
& CC-BY-SA-4.0. \\

\midrule

\multicolumn{3}{l}{\textit{Zero-shot evaluation}} \\

PTB~\citep{marcus1993building}
& Zero-shot evaluation
& LDC User Agreement for Non-Members. \\

WikiText~\citep{merity2016pointer}
& Zero-shot evaluation
& CC-BY-SA-4.0. \\

LAMBADA~\citep{paperno2016lambada}
& Zero-shot evaluation
& CC-BY-4.0. \\

AG News~\citep{zhang2015character}
& Zero-shot evaluation
& Research / non-commercial use. \\

Scientific Papers~\citep{cohan2018discourse}
& Zero-shot evaluation
& Source-specific OA licenses; repository code under Apache-2.0. \\

\bottomrule
\end{tabular}
\end{table}

\clearpage
\newpage
\end{document}